\documentclass[11pt,a4paper]{article}
\usepackage[hyperref]{acl2021}
\usepackage{times}
\usepackage{latexsym}

\usepackage{makecell}
\usepackage[ruled]{algorithm2e}
\usepackage{amsthm}
\usepackage{amsmath}
\usepackage{amsfonts}
\usepackage{multirow}
\usepackage{enumitem}
\usepackage{array}
\usepackage{mathrsfs}
\usepackage{stmaryrd} 
\usepackage{subfigure}
\usepackage{balance}
\usepackage{graphicx}
\usepackage{amsfonts}
\usepackage{url}
\usepackage{xspace}
\usepackage{upgreek}
\usepackage{color,xcolor}
\usepackage{amsmath}
\usepackage{makecell}
\usepackage{multirow}
\usepackage{array}
\usepackage{booktabs}
\usepackage{amssymb}
\usepackage{amsmath}
\usepackage{amsfonts}
\usepackage{multirow}
\usepackage{booktabs}

\definecolor{lavender}{rgb}{0.9, 0.9, 0.98}

\usepackage{colortbl}
\newenvironment{noindlist}{
\begin{list}{\labelitemi}{
\leftmargin=1.0em 
\itemindent=0em 
\itemsep=2pt 
\parsep=1pt 
\parskip=1pt
}}{\end{list}}

\usepackage{cleveref}
\crefformat{section}{\S#2#1#3}
\crefformat{subsection}{\S#2#1#3}
\crefformat{subsubsection}{\S#2#1#3}
\crefrangeformat{section}{\S\S#3#1#4 to~#5#2#6}
\crefmultiformat{section}{\S\S#2#1#3}{ and~#2#1#3}{, #2#1#3}{ and~#2#1#3}
\usepackage{refstyle}
\Crefformat{figure}{#2Fig.~#1#3}
\Crefmultiformat{figure}{Figs.~#2#1#3}{ and~#2#1#3}{, #2#1#3}{ and~#2#1#3}
\Crefformat{table}{#2Tab.~#1#3}
\Crefmultiformat{table}{Tabs.~#2#1#3}{ and~#2#1#3}{, #2#1#3}{ and~#2#1#3}
\Crefformat{appendix}{Appx.~\S#2#1#3}
\crefformat{algorithm}{Alg.~#2#1#3}

\usepackage{microtype}

\aclfinalcopy % Uncomment this line for the final submission
\newcommand{\stitle}[1]{\vspace{1ex} \noindent{\bf #1.}}

\newcommand{\dataset}{DBP2.0\xspace}

\title{Knowing the No-match: Entity Alignment with Dangling Cases}

\author{Zequn Sun$^1$ \quad Muhao Chen$^{2,3}$ \quad Wei Hu$^1$\\
	$^1$State Key Laboratory for Novel Software Technology, Nanjing University, China \\
	$^2$Department of Computer Science, University of Southern California, USA \\
	$^3$Information Sciences Institute, University of Southern California, USA \\
	\normalsize\texttt{zqsun.nju@gmail.com,\,muhaoche@usc.edu,\,whu@nju.edu.cn} \\
}

\date{}

\begin{document}
\maketitle
\begin{abstract}
This paper studies a new problem setting of entity alignment for knowledge graphs (KGs). Since KGs possess different sets of entities, there could be entities that cannot find alignment across them, leading to the problem of \emph{dangling entities}. As the first attempt to this problem, we construct a new dataset and design a multi-task learning framework for both entity alignment and dangling entity detection. The framework can opt to abstain from predicting alignment for the detected dangling entities. We propose three techniques for dangling entity detection that are based on the distribution of nearest-neighbor distances, i.e., nearest neighbor classification, marginal ranking and background ranking. After detecting and removing dangling entities, an incorporated entity alignment model in our framework can provide more robust alignment for remaining entities. Comprehensive experiments and analyses demonstrate the effectiveness of our framework. We further discover that the dangling entity detection module can, in turn, improve alignment learning and the final performance. The contributed resource is publicly available to foster further research.
\end{abstract}

\section{Introduction}
Knowledge graphs (KGs) have evolved to be the building blocks of many intelligent systems~\cite{KG_survey}.
Despite the importance, KGs are usually costly to construct~\cite{paulheim2018much} and naturally suffer from incompleteness~\cite{KBCompleteness}. 
Hence, merging multiple KGs through entity alignment can lead to mutual enrichment of their knowledge \cite{KENS}, and provide downstream applications with more comprehensive knowledge representations~\cite{LinkNBed,KENS}. Entity alignment seeks to discover identical entities in different KGs, such as English entity \texttt{Thailand} and its French counterpart \texttt{Thaïlande}. 
To tackle this important problem, literature has attempted with the embedding-based entity alignment methods~\cite{MTransE,GCN_Align,MuGNN,GraphMatch_iclr20,NMN_acl20,AttrGNN,HyperKA}. 
These methods jointly embed different KGs and put similar entities at close positions in a vector space, where the nearest neighbor search can retrieve entity alignment.
Due to its effectiveness, embedding-based entity alignment has drawn extensive attention in recent years \cite{OpenEA}.

\begin{figure}[t]
	\centering
	\includegraphics[width=0.96\linewidth]{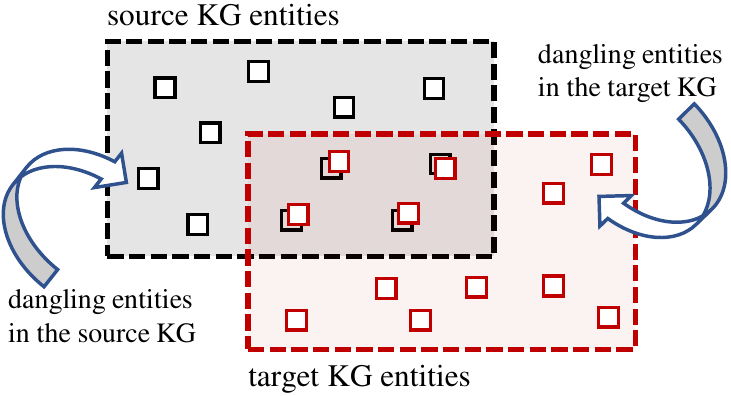}
	\caption{Illustration of entity alignment between two KGs with dangling cases. Paired red and black squares in the overlap region denote entity alignment while others are dangling entities without counterparts.}
	\label{fig:na_ea}
\end{figure}

Nonetheless, to practically support the alignment of KGs as a real-world task, existing studies suffer one common problem of identifying entities without alignment across KGs (called \emph{dangling entities}).
Specifically, current methods are all built upon the assumption that any source entity has a counterpart in the target KG~\cite{OpenEA}, and are accordingly developed with learning resources that enforce the same assumption.
Hence, given every entity in a source KG, a model always tends to predict a counterpart via the nearest neighbor search in the embedding space.
However, since each KG may be independently created based on separate corpora~\cite{DBpedia} or contributed by different crowds~\cite{speer2017conceptnet,carlson2010toward},
it is natural for KGs to possess different sets of entities \cite{Integration}, as illustrated in \Cref{fig:na_ea}. 
Essentially, this problem overlooked in prior studies causes existing methods to fall short of distinguishing between matchable and dangling entities,
hence hinders any of such methods to align KGs in a real-world scenario.

Towards more practical solutions of entity alignment for KGs,
we provide a redefinition of the task with the incorporation of dangling cases (\Cref{sect:task}), as the \textit{first contribution} of this work. 
Given a source entity, our setting does not assume that it must have a counterpart in the target KG as what previous studies do. 
Instead, conducting entity alignment also involves identifying whether the counterpart of an entity actually exists in another KG.
Hence, a system to tackle this realistic problem setting of entity alignment is also challenged by the requirement for justifying the validity of its prediction.

To facilitate the research towards the new problem,
the \textit{second contribution} of this work is to construct a new dataset \dataset for entity alignment with dangling cases (\Cref{sect:dataset}). 
As being discussed, existing benchmarks for entity alignment, including DBP15K~\cite{JAPE}, WK3L~\cite{MTransE} and the more recent OpenEA~\cite{OpenEA}, are set with the constraint that any entity to be aligned should have a valid counterpart.
We use the full DBpedia~\cite{DBpedia} to build a new dataset and the key challenge lies in that we need to guarantee the selected dangling entities actually do not have counterparts. 
We first extract two subgraphs with one-to-one entity alignment (i.e., all entities have counterparts). 
Then, we randomly remove some entities to make their left counterparts in the peer KG dangling.

Although embedding-based entity alignment has been investigated for several years, handling with dangling entities has not been studied yet. 
As the \textit{third contribution}, we present a multi-task learning framework for the proposed task (\Cref{sect:model}). 
It consists of two jointly optimized modules for \emph{entity alignment} and \emph{dangling entity detection}, respectively. 
While the entity alignment module can basically incorporate any existing techniques from prior studies~\cite{OpenEA},
in this paper, we experiment with two representative techniques, i.e., relational embedding based~\cite{MTransE} and neighborhood aggregation based~\cite{AliNet} methods.
For dangling entity detection, our framework incorporates an auxiliary learning objective, which seeks to learn a confidence metric for the inferred entity alignment.
The principle to realize such metric learning is that the embeddings of dangling entities should be isolated and are distant from others. 
According to this principle, we exploit several techniques to distinguish between matchable and dangling entities based on their distance distribution with their neighbors (\Cref{sect:model}), including nearest neighbor classification, marginal ranking and background ranking~\cite{Network_Agnostophobia}.

We conduct comprehensive experiments on the new \dataset dataset, which demonstrate the proposed techniques to solve the dangling entity detection problem to different extents.
Moreover, we observe that training the dangling detection model (marginal ranking) provides an effective indirect supervision that improves the detection of alignment for matchable entities. 
We hope our task, dataset and framework can foster further investigation of entity alignment techniques in the suggested real scenario, leading to more effective and practical solutions to this challenging but important problem.

\section{Task and Dataset}
We hereby describe the problem setting of our task and introduce the new dataset.

\subsection{Task Definition}\label{sect:task}
A KG is a set of relational triples  $\mathcal{T} \subseteq \mathcal{E}\times \mathcal{R}\times\mathcal{E}$, where $\mathcal{E}$ and $\mathcal{R}$ denote vocabularies of entities and relations, respectively. Without loss of generality, we consider entity alignment between two KGs, i.e., a source KG $\mathcal{K}_1\!=\!(\mathcal{T}_1, \mathcal{E}_1, \mathcal{R}_1)$ and a target KG $\mathcal{K}_2\!=\!(\mathcal{T}_2,\mathcal{E}_2, \mathcal{R}_2)$. Given a small set of seed entity alignment $\mathcal{A}_{12}=\{(e_1, e_2) \in \mathcal{E}_1\times\mathcal{E}_2\|e_1\equiv e_2\}$ along with a small set of source entities $\mathcal{D}\subset\mathcal{E}_{1}$ known to have no counterparts as training data, 
the task seeks to find the remaining entity alignment.
Different from the conventional entity alignment setting \cite{JAPE},
a portion (with an anticipated quantity) of entities in $\mathcal{E}_1$ and $\mathcal{E}_2$ may have no counterparts.
Our training and inference stages take such dangling entities into consideration.

\subsection{Dataset Construction}\label{sect:dataset}

As discussed, previous testbeds for entity alignment do not contain dangling entities \cite{JAPE,KDCoE,OpenEA}.
Therefore, we first create a new dataset to support the study of the proposed problem setting.
Same as the widely used existing benchmark DBP15K \cite{JAPE}, we choose DBpedia 2016-10\footnote{Downloaded from \url{https://wiki.dbpedia.org/downloads-2016-10}. The latest 2020 version has not provided updated data for some languages other than English when this study is conducted.} as the raw data source.
Following DBP15K, we also use English (EN), French (FR), Japanese (JA) and Chinese (ZH) versions of DBpedia to build three entity alignment settings of ZH-EN, JA-EN and FR-EN.
For each monolingual KG, the triples are extracted from the Infobox Data of DBpedia, where relations are not mapped to a unified ontology. 
The reference entity alignment data is from the inter-language links (ILLs) of DBpedia across these three bridges of languages.
Such reference data is later used as alignment labels for training and testing, and also serves as references to recognize dangling entities.

\begin{table}[!t]	
	\centering
% 	\resizebox{0.99\linewidth}{!}{
	{\small
	\setlength{\tabcolsep}{4pt}
		\begin{tabular}{clrrrr}
			\toprule			\multicolumn{2}{c}{Datasets} & \# Entities & \# Rel. & \# Triples & \# Align. \\ \midrule
			\multirow{2}{*}{ZH-EN} 
			& ZH & 84,996 & 3,706 & 286,067 &\multirow{2}{*}{33,183} \\
			& EN & 118,996 & 3,402 & 586,868 & \\ \midrule
			\multirow{2}{*}{JA-EN} 
			& JA & 100,860 & 3,243 & 347,204 &\multirow{2}{*}{39,770} \\
			& EN & 139,304 & 3,396 & 668,341 & \\ \midrule
			\multirow{2}{*}{FR-EN}
			& FR & 221,327 & 2,841 & 802,678 &\multirow{2}{*}{123,952} \\
			& EN & 278,411 & 4,598 & 1,287,231 & \\
			\bottomrule
	\end{tabular}}
% 	}
	\caption{\label{tab:dataset}Statistics of the \dataset dataset.}
\end{table}

\stitle{Construction} 
The key challenge of building our dataset lies in that we need to ensure the selected dangling entities are indeed without counterparts. 
Specifcally, we cannot simply regard entities without ILLs as dangling ones, since the ILLs are also incomplete \cite{MTransE}. 
Under this circumstance, we use a two-step dataset extraction process, which first samples two subgraphs whose entities all have counterparts based on ILLs, 
and randomly removes a disjoint set of entities in the source and target graphs to make their counterparts dangling. 
For the first step, we iteratively delete unlinked entities and their triples from the source and target KGs until the left two subgraphs are one-to-one aligned. 
In the second step for entity removal, while the removed entities are disjoint in two KGs, the proportion of the removed entities 
also complies with the proportion of unaligned entities in each KG.

\stitle{Statistics and evaluation} 
\Cref{tab:dataset} lists the statistics our dataset. 
The three entity alignment settings have different data scales and each is much larger than the same setting in DBP15K, 
thus can benefit better scalability analysis of models.
For dangling entity detection, we split $30\%$ of dangling entities for training, $20\%$ for validation and others for testing. 
The splits of reference alignment follow the same partition ratio,
which is also consistent with that of DBP15K to simulate the weak alignment nature of KGs \cite{MTransE,JAPE}. 
We also compare the degree distribution of matchable and dangling entities in our dataset against DBP15K in \Cref{fig:degree} of \Cref{appendix:degree}. 
We find the matchable and unlabeled entities in DBP15K have biased degree distribution, which has an adverse effect on dangling entity detection and leads to unreal evaluation. By contrast, in \dataset, matchable and dangling entities have similar degree distribution.
\section{Entity Alignment with Dangling Cases}
\label{sect:model}

\begin{figure}[t]
	\centering
	\includegraphics[width=0.999\linewidth]{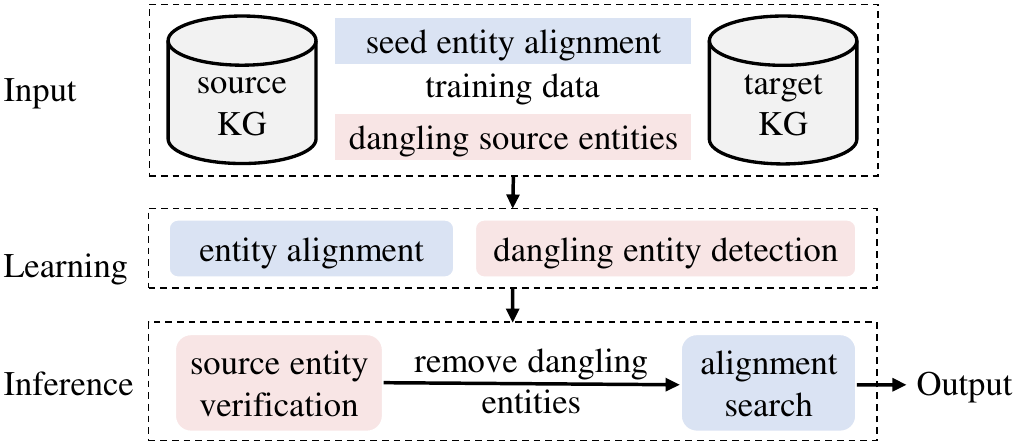}
	\caption{Framework of entity alignment w/ abstention.}
	\label{fig:framework}
\end{figure}

We propose a multi-task learning framework for entity alignment with dangling cases, as illustrated in \Cref{fig:framework}. 
It has two jointly optimized modules, i.e., entity alignment and dangling entity detection. 
The entity alignment module takes as input relational triples of two KGs (for KG embedding) and seed entity alignment (for alignment learning). 
As for the detection of dangling entities, the module uses a small number of labeled dangling entities
to jump-start the learning of a confidence metric for distinguishing between matchable and dangling entities. 
In the inference stage for entity alignment, 
our framework is able to first identify and remove dangling entities, then predict alignment for those that are decided to be matchable. 

\subsection{Entity Alignment}
Our framework can incorporate any entity alignment technique.
For the sake of generality, we consider two representative techniques in our framework.
One technique is based on MTransE \cite{MTransE}, which is among the earliest studies for embedding-based entity alignment. 
It employs the translational model TransE \cite{TransE} to embed KGs in separate spaces, meanwhile jointly learns a linear transformation
between the embedding spaces to match entity counterparts.
Specifically, given an entity pair $(x_1,x_2)\in\mathcal{A}_{12}$, let $\mathbf{x}_1$ and $\mathbf{x}_2$ be their embeddings learned by the translational model. 
MTransE learns the linear transformation induced by a matrix $\mathbf{M}$ by minimizing $\|\mathbf{M}\mathbf{x}_1\!-\!\mathbf{x}_2\|$, where $\|\!\cdot\!\|$ denotes the $L_1$ or $L_2$ norm.

The other technique is from AliNet \cite{AliNet}, which is one of the SOTA methods based on graph neural networks. 
AliNet encodes entities by performing a multi-hop neighborhood aggregation, seeking to cope with heteromorphism of their neighborhood structures. 
For alignment learning, different from MTransE that only minimizes the transformed embedding distance, AliNet additionally optimizes a margin-based ranking loss
for entity counterparts with negative samples. 
Specifically, let $x$ be a matchable source entity in the seed entity alignment, and $x'$ is a randomly-sampled entity in the target KG, AliNet attempts to ensure $\|\mathbf{x}-\mathbf{x}'\|>\lambda_1>0$, where $\lambda_1$ is a distance margin. 

\subsection{Dangling Entity Detection}\label{sec:dangling}
We propose three techniques to implement the dangling detection module based on the distribution of the nearest neighbor distance in embedding space.

\subsubsection{NN Classification}
This technique is to train a binary classifier to distinguish between dangling entities (labeled $1$, i.e., $y=1$) and matchable ones ($y=0$). 
Specifically, we experiment with a feed-forward network (FFN) classifier.
Given a source entity $x$, its input feature representation is the difference vector between its embedding $\mathbf{x}$ and its transformed NN embedding $\mathbf{x}_\text{nn}$ in the target KG embedding space\footnote{We use \emph{transformed nearest neighbor (NN)} to denote the the NN of a source KG entity after it is transformed to the target embedding space.}. 
The confidence of $x$ being a dangling entity is given by $p(y=1|x) = \text{sigmoid}(\text{FFN}(\mathbf{M}\mathbf{x}-\mathbf{x}_\text{nn}))$. 
Let $\mathcal{D}$ be the training set of dangling source entities and $\mathcal{A}$ denotes the set of matchable entities in the training alignment data. 
For every $x\in\mathcal{D}\cup\mathcal{A}$, we minimize the cross-entropy loss:
\begin{equation}\label{eq:cross-entropy}
\begin{split}
\mathcal{L}_x = - \big(& y_x\log(p(y=1|x)) \\ &+ (1-y_x)\log(1-p(y=1|x))\big),
\end{split}
\end{equation}
where $y_x$ denotes the truth label for entity $x$. In a real-world entity alignment scenario, the dangling entities and matchable ones usually differ greatly in quantity, leading to unbalanced label distribution.
In that case, we apply label weights \cite{huang2016learning} to balance between the losses for both labels.

\subsubsection{Marginal Ranking}
Considering that dangling entities are the noises for finding entity alignment based on embedding distance, we are motivated to let dangling entities have solitary representations in the embedding space, i.e., they should keep a distance away from their surrounding embeddings. 
Hence, we seek to put a distance margin between dangling entities and their sampled NNs. 
For every input dangling entity $x\in\mathcal{D}$, we minimize the following loss:
\begin{equation}\label{eq:margin}
\mathcal{L}_x = \max(0, \lambda - \|\mathbf{M}\mathbf{x}-\mathbf{x}_\text{nn}\|),
\end{equation}
where $\lambda$ is a distance margin. 
This loss and the entity alignment loss (e.g., that of MTransE) conduct joint learning-to-rank, i.e., the distance between unaligned entities should be larger than that of aligned entities while dangling entities should have a lower ranking in the candidate list of any source entity.

\subsubsection{Background Ranking}
In the two aforementioned techniques,
searching for the NN of an entity is time-consuming.
Furthermore, selecting an appropriate value for the distance margin of the second technique is not trivial. 
Based on empirical studies, we find that the margin has a significant influence on the final performance.
Hence, we would like to find a more efficient and self-driven technique. 
Inspired by the open-set classification approach \cite{Network_Agnostophobia} that lets a classifier equally penalize the output logits for samples of classes that are unknown to training (i.e. \emph{background classes}), 
we follow a similar principle and let the model equally enlarge the distance of a dangling entity from any sampled target-space entities.
This method is to treat all dangling entities as the ``background" of the embedding space, since they should be distant from matchable ones.
We also decrease the scale of the dangling entity embeddings to further provide a separation between the embeddings of matchable and dangling entities.
For the dangling entity $x\in\mathcal{D}$, let $X^v_x$ be the set of randomly-sampled target entities with size of $v$. The loss is defined as
\begin{equation}\label{eq:background}
\mathcal{L}_x = \sum_{x'\in X^v_x} \big|\lambda_x - \|\mathbf{M}\mathbf{x}-\mathbf{x}'\|\big| + \alpha\|\mathbf{x}\|,
\end{equation}
where $|\cdot|$ denotes the absolute value and $\alpha$ is a weight hyper-parameter for balance. $\lambda_x$ is the average distance, i.e., $\lambda_x = \frac{1}{v} \sum_{x'\in X^v_x} \| \mathbf{M}\mathbf{x}-\mathbf{x}'\|$. 
This objective can push the relatively close entities away from the source entity without requiring a pre-defined distance margin.

\subsection{Learning and Inference}
The overall learning objective of the proposed framework is a combination of the entity alignment loss (e.g., MTransE's loss) and one of the dangling entity detection loss as mentioned above. 
The two losses are optimized in alternate batches.
More training details are presented in \Cref{sec:config}.

Like the training phase,
the inference phase is also separated into dangling entity detection and entity alignment.
The way of inference for dangling entities differs with the employed technique.
The NN classification uses the jointly trained FFN classifier to estimate whether the input entity is a dangling one.
The marginal ranking takes the preset margin value in training as a confidence threshold, and decides whether an entity is a dangling one based on if its transformed NN distance is higher than the threshold.
The inference of background ranking is similar to that of marginal ranking, with only the difference, by its design, to be that the confidence threshold is set as the average NN distance of entities in the target embedding space.
After detecting dangling entities, the framework finds alignment in the remaining entities based on the transformed NN search among the matchable entities in the embedding space of the target KG.

\stitle{Accelerated NN search}
The first and second techniques need to search NNs. We can use an efficient similarity search library Faiss \cite{faiss} for fast NN retrieval in large embedding space. 
We also maintain a cache to store the NNs of entities backstage and update it every ten training epochs.

\section{Experiments}
In this section, we report our experimental results.
We start with describing the experimental setups (\Cref{sec:setting}). Next, we separately present the experimentation under two different evaluation settings (\Cref{sec:relaxed}-\Cref{sec:consolidated}), followed by an analysis on the similarity score distribution of the obtained representations for matchable and dangling entities (\Cref{sect:viz}).
To faciliate the use of the contributed dataset and software, we have incorporated these resources into the OpenEA benchmark\footnote{\url{https://github.com/nju-websoft/OpenEA}}~\cite{OpenEA}.

\begin{table*}[!t]
	\centering
	%\centering\renewcommand\arraystretch{0.9}
	%\resizebox{.999\textwidth}{!}{
	{\small
	\setlength{\tabcolsep}{1pt}
		\begin{tabular}{lcccccccccccccccccc}
			\toprule
			\multirow{2}{*}{Methods} &
			\multicolumn{3}{c}{ZH-EN} & \multicolumn{3}{c}{EN-ZH} & \multicolumn{3}{c}{JA-EN} & \multicolumn{3}{c}{EN-JA} & \multicolumn{3}{c}{FR-EN} & \multicolumn{3}{c}{EN-FR}\\
			\cmidrule(lr){2-4} \cmidrule(lr){5-7} \cmidrule(lr){8-10} \cmidrule(lr){11-13} \cmidrule(lr){14-16} \cmidrule(lr){17-19}
			& H@1 & H@10 & MRR & H@1 & H@10 & MRR & H@1 & H@10 & MRR & H@1 & H@10 & MRR & H@1 & H@10 & MRR & H@1 & H@10 & MRR \\ 
			\midrule
			MTransE & {.358} & {.675} & {.463} & .353 & {.670} & {.461} & {.348} & {.661} & {.453} & .342 & {.670} & .452 & {.245} & {.524} & {.338} & {.247} & {.531} & {.342} \\
% 			\rowcolor{lavender}
		   \;\;w/ NNC & .350 & .668 & .457 & .356 & .664 & .460 & .340 & .657 & .441 & .336 & .630 & .445 & .253 & .539 & .343 & .251 & .536 & .343 \\
		  % \rowcolor{lavender}
			\;\;w/ MR & \textbf{.378} & \textbf{.693} & \textbf{.487} & \textbf{.383}& \textbf{.699} & \textbf{.491} & \textbf{.373} & \textbf{.686} & \textbf{.476} & \textbf{.374} & \textbf{.707} & .\textbf{485} & \textbf{.259} & \textbf{.541} & \textbf{.348} & \textbf{.265} & \textbf{.553} & \textbf{.360} \\
% 			\rowcolor{lavender}
			\;\;w/ BR & .360 & .678 & .468 & .357 & .675 & .465 & .344 & .660 & .451 & .346 & .675 & .456 & .251 & .525 & .342 & .249 & .531 & .343 \\
			\midrule
			AliNet & .332 & .594 & .421 & {.359} & .629 & .451 & .338 & .596 & .429 & {.363} & .630 & {.455} & .223 & .473 & .306 & .246 & .495 & .329 \\
% 			\rowcolor{lavender}
			\;\;w/ NNC & .321 & .598 & .415 & .335 & .608 & .428 & .330 & .602 & .422 & .344 & .627 & .439 & .212 & .467 & .294 & .230 & .476 & .312 \\
% 			\rowcolor{lavender}
			\;\;w/ MR & \textbf{.343} & \textbf{.606} & \textbf{.433} & \textbf{.364} & \textbf{.637} & \textbf{.459} & \textbf{.349} & \textbf{.608} & \textbf{.438} & \textbf{.377} & \textbf{.646} & \textbf{.469} & \textbf{.230} & \textbf{.477} & \textbf{.312} & \textbf{.252} & \textbf{.502} & \textbf{.335}\\
% 			\rowcolor{lavender}
			\;\;w/ BR & .333 & .599 & .426 & .357 & .632 & .451 & .341 & \textbf{.608} & .431 & .369 & .636 & .461 & .214 & .468 & .298 & .238 & .487 & .321 \\
			\bottomrule
	\end{tabular}}
	\caption{Entity alignment results (relaxed setting) of MTransE and AliNet on \dataset.}
	\label{tab:synthetic_ent_alignment}
\end{table*}

\subsection{Experimental Settings}\label{sec:setting}
We consider two evaluation settings. 
One setting is for the proposed problem setting with dangling entities, for which we refer as the \emph{consolidated evaluation setting}. 
We first detect and remove the dangling source entities and then search alignment for the left entities.
For this evaluation setting, we also separately assess the performance of the dangling detection module.
The other simplified setting follows that in previous studies \cite{JAPE,OpenEA} where the source entities in test set all have counterparts in the target KG, so no dangling source entities are considered. 
In this \emph{relaxed evaluation setting}, we seek to evaluate the effect of dangling entity detection on entity alignment and make our results comparable to previous work. 

\stitle{Evaluation Protocol}
For the \emph{relaxed evaluation setting}, given each source entity, the candidate counterpart list is selected via NN search in the embedding space. 
The widely-used metrics on the ranking lists are Hits@$k$ ($k=1,10$, H@$k$ for short) and mean reciprocal rank (MRR).
Higher H@$k$ and MRR indicate better performance.

For the \emph{consolidated setting},
we report precision, recall and F1 for dangling entity detection.
As for assessing the eventual performance of realistic entity alignment, since the dangling entity detection may not be perfect,
it is inevitable for some dangling entities to be incorrectly sent to the entity alignment module for aligning, while some matchable ones may be wrongly excluded.
In this case, H@$k$ and MRR are not applicable for the consolidated entity alignment evaluation. 
Following a relevant evaluation setting for entity resolution in database \cite{DL4ER,EmbedER}, we also use precision, recall and F1 as metrics.
More specifically, if a source entity is dangling and is not identified by the detection module, the prediction is always regarded as incorrect. 
Similarly, if a matchable entity is falsely excluded by the dangling detection module, this test case is also regarded as incorrect since the alignment model has no chance to search for alignment. 
Otherwise, the alignment module searches for the NN of a source entity in the target embedding space and assesses if the predicated counterpart is correct. 

\begin{figure}[t]
	\centering
	\includegraphics[width=.96\linewidth]{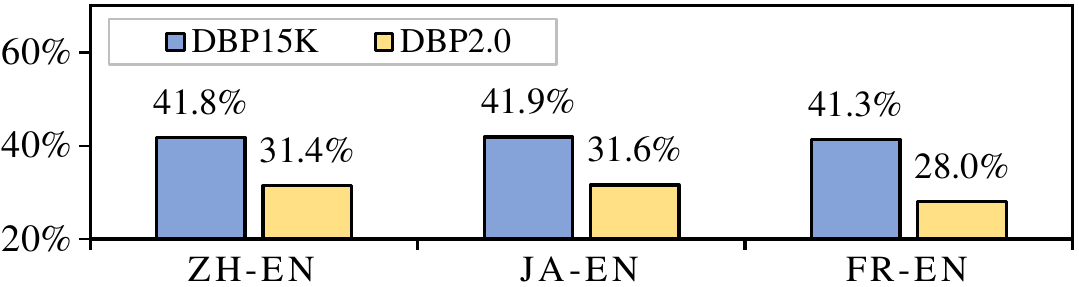}
	\caption{Average neighbor overlap ratio of aligned entities in DBP15K and our \dataset.}
	\label{fig:data_comp}
\end{figure}

\begin{table*}[!t]
	\centering
	%\renewcommand\arraystretch{0.9}
	%\resizebox{.999\textwidth}{!}{\small
	{
	\small
	\setlength{\tabcolsep}{3pt}
		\begin{tabular}{llcccccccccccccccccc}
			\toprule
% 			\multirow{2}{*}{Methods} & 
            \multicolumn{2}{c}{\multirow{2}{*}{Methods}} &
			\multicolumn{3}{c}{ZH-EN} & \multicolumn{3}{c}{EN-ZH} & \multicolumn{3}{c}{JA-EN} & \multicolumn{3}{c}{EN-JA} &  \multicolumn{3}{c}{FR-EN} & \multicolumn{3}{c}{EN-FR}\\
			\cmidrule(lr){3-5} \cmidrule(lr){6-8} \cmidrule(lr){9-11} \cmidrule(lr){12-14} \cmidrule(lr){15-17} \cmidrule(lr){18-20}
			&& Prec. & Rec. & F1 & Prec. & Rec. & F1 & Prec. & Rec. & F1 & Prec. & Rec. & F1 & Prec. & Rec. & F1 & Prec. & Rec. & F1 \\ 
			\midrule
			\parbox[t]{2mm}{\multirow{3}{*}{\rotatebox[origin=c]{90}{{\scriptsize MTransE}}}} 
			& NNC & .604 & .485 & .538 & .719 & .511 & .598 & .622 & .491 & .549 & .686 & .506 & .583 & .459 & .447 & .453 & .557 & .543 & .550 \\
			& MR & .781 & .702 & .740 & .866 & .675 & .759 & .799 & .708 & .751 & .864 & .653 & .744 & .482 & .575 & .524 & .639 & .613 & .625 \\
			& BR & \textbf{.811} & \textbf{.728} & \textbf{.767} & .\textbf{892} & \textbf{.700} & \textbf{.785} & \textbf{.816} & \textbf{.733} & \textbf{.772} & \textbf{.888} & \textbf{.731} & \textbf{.801} & \textbf{.539} & \textbf{.686} & \textbf{.604} & \textbf{.692} & \textbf{.735} & \textbf{.713} \\
			\midrule
			\parbox[t]{2mm}{\multirow{3}{*}{\rotatebox[origin=c]{90}{{\small AliNet}}}}
			& NNC & .676 & .419 & .517 & .738 & .558 & .634 & .597 & .482 & .534 & .761 & .120 & .207 & .466 & .365 & .409 & .545 & .162 & .250 \\
			& MR & .752 & .538 & .627 & .828 & .505 & .627 & .779 & .580 & .665 & \textbf{.854} & .543 & \textbf{.664} & \textbf{.552} & \textbf{.570} & \textbf{.561} & \textbf{.686} & .549 & \textbf{.609} \\
			& BR & \textbf{.762} & \textbf{.556} & \textbf{.643} & \textbf{.829} & \textbf{.515} & \textbf{.635} & \textbf{.783} & \textbf{.591} & \textbf{.673} & .846 & \textbf{.546} & .663 & .547 & .556 & .552 & .674 & \textbf{.556} & \textbf{.609} \\
			\bottomrule
	\end{tabular}}
	\caption{Dangling entity detection results on \dataset.}
	\label{tab:detection}
\end{table*}

\stitle{Model Configuration}\label{sec:config}
As described in \Cref{sec:dangling}, our dangling detection module has three variants, i.e., NN classification (NNC), marginal ranking (MR), and background ranking (BR). 
We report the implementation details of the entity alignment module (w/ MTransE or AliNet) in \Cref{appendix:config,appendix:setting}.
We initialize KG embeddings and model parameters using the Xavier initializer \cite{Xavier}, and use Adam \cite{Adam} to optimize the learning objectives with the learning rate $0.001$ for MTransE and $0.0005$ for AliNet. 
Note that we do not follow some methods to initialize with machine translated entity name embeddings \cite{NMN_acl20}.
As being pointed out by recent studies \cite{JEANS,EVA,AttrGNN}, this is necessary to prevent test data leakage.
Entity similarity is measured by cross-domain similarity local scaling \cite{CSLS} for reduced hubness effects, as being consistent to recent studies \cite{AliNet,JEANS}.
We use a two-layer FFN in NNC. 
For MR, the margin is set as $\lambda=0.9$ for MTransE and $0.2$ for AliNet.
BR randomly samples $20$ target entities for each entity per epoch and $\alpha=0.01$.
Training is terminated based on F1 results of entity alignment on validation data.

\begin{table*}[!t]
	\centering
	%\renewcommand\arraystretch{0.9}
	%\resizebox{.99\textwidth}{!}{\small
	{
	\small
	\setlength{\tabcolsep}{3pt}
		\begin{tabular}{llcccccccccccccccccc}
			\toprule
			\multicolumn{2}{c}{\multirow{2}{*}{Methods}} &
			\multicolumn{3}{c}{ZH-EN} & \multicolumn{3}{c}{EN-ZH} & \multicolumn{3}{c}{JA-EN} & \multicolumn{3}{c}{EN-JA} &  \multicolumn{3}{c}{FR-EN} & \multicolumn{3}{c}{EN-FR}\\
			\cmidrule(lr){3-5} \cmidrule(lr){6-8} \cmidrule(lr){9-11} \cmidrule(lr){12-14} \cmidrule(lr){15-17} \cmidrule(lr){18-20}
			&& Prec. & Rec. & F1 & Prec. & Rec. & F1 & Prec. & Rec. & F1 & Prec. & Rec. & F1 & Prec. & Rec. & F1 & Prec. & Rec. & F1 \\ 
			\midrule
			\parbox[t]{2mm}{\multirow{3}{*}{\rotatebox[origin=c]{90}{{\scriptsize MTransE}}}} 
			& NNC & .164 & .215 & .186 & .118 & .207 & .150 & .180 & .238 & .205 & .101 & .167 & .125 & .185 & .189 & .187 & .135 & .140 & .138 \\
			& MR & .302 & .349 & .324 & .231 & .362 & .282 & .313 & \textbf{.367} & \textbf{.338} & .227 & \textbf{.366} & .280 & .260 & \textbf{.220} & \textbf{.238} & .213 & \textbf{.224} & .218 \\
			& BR & \textbf{.312} & \textbf{.362} & \textbf{.335} & \textbf{.241} & \textbf{.376} & \textbf{.294} & \textbf{.314} & .363 & .336 & \textbf{.251} & .358 & \textbf{.295} & \textbf{.265} & .208 & .233 & \textbf{.231} & .213 & \textbf{.222} \\
			\midrule
			\parbox[t]{2mm}{\multirow{3}{*}{\rotatebox[origin=c]{90}{{\small AliNet}}}} 
			& NNC & .121 & .193 & .149 & .085 & .138 & .105 & .113 & .146 & .127 & .067 & .208 & .101 & .126 & .148 & .136 & .086 & .161 & .112 \\
			& MR & \textbf{.207} & \textbf{.299} & \textbf{.245} & \textbf{.159} & \textbf{.320} & \textbf{.213} & \textbf{.231} & \textbf{.321} & \textbf{.269} & \textbf{.178} & \textbf{.340} & \textbf{.234} & \textbf{.195} & \textbf{.190} & \textbf{.193} & .160 & \textbf{.200} & .178 \\
			& BR & .203 & .286 & .238 & .155 & .308 & .207 & .223 & .306 & .258 & .170 & .321 & .222 & .183 & .181 & .182 & \textbf{.164} & \textbf{.200} & \textbf{.180} \\
			\bottomrule
	\end{tabular}}
	\caption{Entity alignment results on \dataset.}
	\label{tab:ent_alignment}
\end{table*}

\subsection{Relaxed Evaluation}\label{sec:relaxed}
We first present the evaluation under the relaxed entity alignment setting based on \Cref{tab:synthetic_ent_alignment}.
This setting only involves matchable source entities to test entity alignment,
which is an ideal (but less realistic) scenario similar to prior studies \cite{OpenEA}. 
We also examine if jointly learning to detect dangling entities can indirectly improve alignment.

As observed,
MTransE, even without dangling detection, can achieve promising performance on \dataset. 
The results are even better than those on DBP15K as reported by \citet{JAPE}.
We attribute this phenomenon to the robustness of this simple embedding method and our improved implementation (e.g., more effective negative sampling).
By contrast, although we have tried our best in tuning, the latest GNN-based AliNet falls behind MTransE.
Unlike MTransE that learns entity embeddings from a first-order perspective (i.e., based on triple plausibility scores), AliNet represents an entity from a high-order perspective by aggregating its neighbor embeddings, and entities with similar neighborhood structures would have similar representations.
However, the dangling entities in \dataset inevitably become spread noises in entity neighborhoods.
To further probe into this issue, we count the average neighbor overlap ratio of aligned entities in DBP15K and our \dataset. Given an entity alignment pair ($x_1,x_2$), let $\pi(x_1)$ and $\pi(x_2)$ be the sets of their neighboring entities
respectively, where we also merge their aligned neighbors as one identity based on reference entity alignment. 
Then the neighbor overlap ratio of $x_1$ and $x_2$ is calculated as $|\pi(x_1)\cap \pi(x_2)|/|\pi(x_1)\cup \pi(x_2)|$. 
We average such a ratio for both DBP15K and \dataset as given in \Cref{fig:data_comp}.
We can see that the three settings' overlap ratios in \dataset are all much lower than those in DBP15K.
Thus, \dataset poses additional challenges, as compared to DBP15K, specifically for those methods relying on neighborhood aggregation.
Based on results and analysis, we argue that methods performing well on the previous synthetic entity alignment dataset may not robustly generalize to the more realistic dataset with dangling cases.
The performance of both MTransE and AliNet is relatively worse on FR-EN, 
which has more entities (i.e., larger candidate search space) and a low neighborhood overlap ratio (therefore, more difficult to match entities based on neighborhood similarity).

Meanwhile,
we find that the dangling detection module can affect the performance of entity alignment.
In details, MR consistently leads to  improvement to both MTransE and AliNet. 
BR can also noticeably boost entity alignment on most settings.
This shows that learning to isolate dangling entities from matchable ones naturally provides indirect help to discriminate the counterpart of a matchable entity from irrelevant ones.
On the other hand, such indirect supervision signals may be consumed by the additional trainable parameters in NNC, causing its effect on entity alignment to be negligible. 
Overall, the observation here calls for more robust entity alignment methods and dangling detection techniques, and lead to further analysis (\Cref{sec:consolidated}).

\begin{figure}[t]
	\centering
	\includegraphics[width=.96\linewidth]{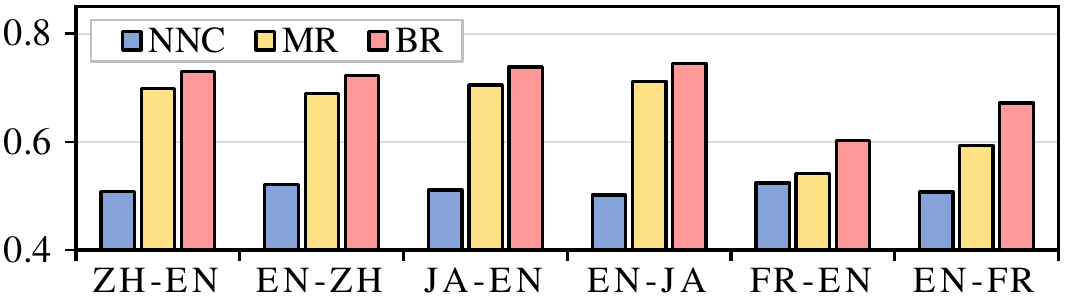}
	\caption{Accuracy of dangling entity detection.}
	\label{fig:accuracy}
\end{figure}

\begin{figure}[t]
	\centering
	\includegraphics[width=.96\linewidth]{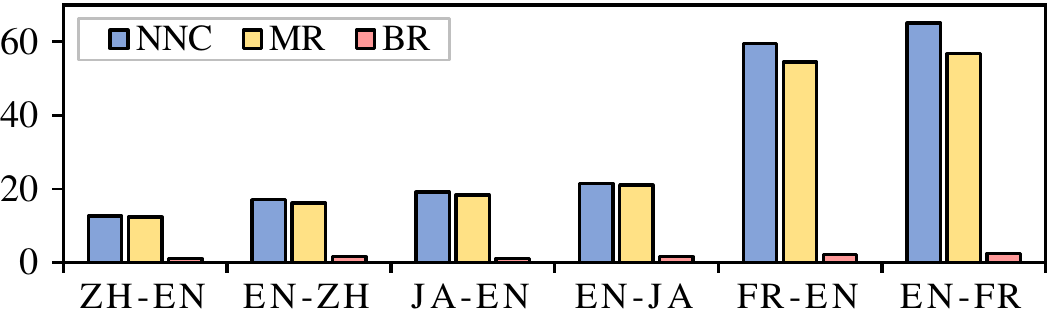}
	\caption{Average training time (seconds) of one epoch for dangling entity detection (MTransE variants).}
	\label{fig:time}
\end{figure}

\subsection{Consolidated Evaluation}\label{sec:consolidated}
We now report the experiment on the more realistic consolidated evaluation setting.
\Cref{tab:detection} gives the precision, recall and F1 results of dangling entity detection, and the final entity alignment performance is presented in \Cref{tab:ent_alignment}. 
In addition, \Cref{fig:accuracy} shows the accuracy of dangling entity detection.
We analyze the results from the following aspects.

\stitle{Dangling entity detection} 
Regardless of which alignment module is incorporated,
NNC performs the worst (e.g., the low recall and accuracy around $0.5$) 
among the dangling detection techniques, whereas BR generally performs the best. 
NNC determines whether an entity is dangling based on the difference vector of the entity embedding and its NN, instead of directly capturing the embedding distance which is observed to be more important based on the results by the other two techniques.
By directly pushing dangling entities away from their NNs in the embedding space, both MR and BR offer much better performance. 
Besides, BR outperforms MR in most cases. 
By carefully checking their prediction results and the actual distance of NNs, we find that the induced distance margin in BR better discriminates dangling entities from matchable ones than the pre-defined margin.

\stitle{Efficiency} 
We compare the average epoch time of training the three dangling detection modules for MTransE in \Cref{fig:time}. We conduct the experiment using a workstation with an Intel Xeon E5-1620 3.50GHz CPU and a NVIDIA GeForce RTX 2080 Ti GPU. 
Since NNC and MR need to search for NNs of source entities, both techniques spend much more training time 
that is saved by random sampling in BR. Overall, BR is an effective and efficient technique for dangling entity detection.

\stitle{Entity alignment}
Generally, for both MTransE and AliNet variants, MR and BR lead to better entity alignment results than NNC. 
MR and BR obtain higher precision and recall performance on detecting dangling entities as listed in \Cref{tab:detection}, resulting in less noise that enters the entity alignment stage.
By contrast, NNC has a low accuracy and thus introduces many noises.
As BR outperforms MR in dangling detection, it also achieves higher entity alignment results than MR on most settings.
We also notice that MR in a few settings, MR offer comparible or slightly better performance than BR.
This is because MR can enhance the learning of alignment modules (see \Cref{sec:relaxed} for detailed analysis), thus delivering improvement to the final performance. 
MTransE variants generally excels AliNet variants in both entity alignment (see \Cref{tab:synthetic_ent_alignment}) and dangling entity detection (see \Cref{tab:detection}) than AliNet, similar to the observation in \Cref{sec:relaxed}.

\stitle{Alignment direction} 
We find that the alignment direction makes a difference in both dangling entity detection and entity alignment. 
Using EN KG as the source is coupled with easier dangling detection than in other languages,
as the most populated EN KG contributes more dangling entities and triples to training than other KGs.
As for entity alignment, we find the observation to be quite the opposite, as using the EN KG as a source leads to noticeable drops in results.
For example, the precision of MTransE-BR is $0.312$ on ZH-EN, but only $0.241$ on EN-ZH.
This is because the EN KG has a larger portion of dangling entities. 
Although the dangling detection module performs well on the EN KG than on others, there are still much more dangling entities entering the alignment search stage, thus reducing the entity alignment precision.
This observation suggests that choosing the alignment direction from a less populated KG to the more populated EN KG can be a more effective solution.

\begin{figure}[t]
	\centering
	\includegraphics[width=.96\linewidth]{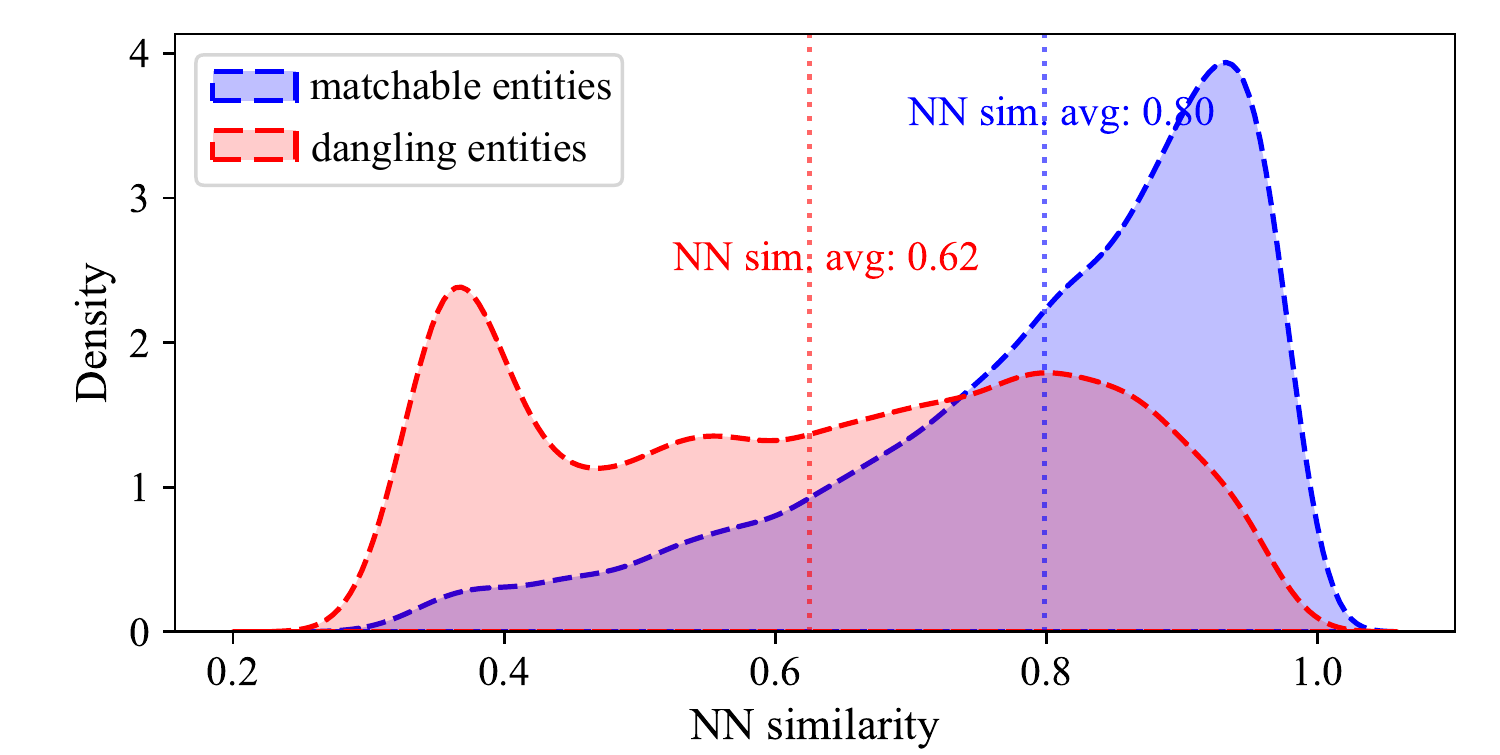}
	\caption{Kernel density estimate plot of the test {\color{blue}{matchable}} and {\color{red}{dangling}} entities' similarity distribution with their nearest target neighbors in ZH-EN.}
	\label{fig:viz}
\end{figure}

\subsection{Similarity Score Distribution}\label{sect:viz}
To illustrate how well the BR technique distinguishes between matchable and dangling entities, we plot in \Cref{fig:viz} the distribution of similarity scores of each test entity and its NN.  
The plot illustrates BR has the expected effect to isolate dangling entities from their NNs, whereas matchable entities are generally placed closer to their NNs.
Yet, we can still see a modest overlap between the two NN similarity distributions of dangling and matchable entities, and a number of dangling entities still have a quite large NN similarity.
This also reveals the fact that the proposed problem setting of entity alignment with dangling cases has many remaining challenges that await further investigation.

\section{Related Work}
We discuss two topics of relevant work.

\subsection{Entity Alignment}
Embedding-based entity alignment is first attempted in MTransE \cite{MTransE}, which jointly learns a translational embedding model and a transform-based alignment model for two KGs.
Later studies generally follow three lines of improvement.
(\romannumeral1) The first line improves the embedding technique to better suit the alignment task, including %TransE-based methods
contextual translation techniques \cite{transedge}, long-term dependency techniques \cite{RSN} and neighborhood aggregation (or GNN-based) ones \cite{GCN_Align,MuGNN,KECG,AliNet,HyperKA,GraphMatch_iclr20}. 
(\romannumeral2) The second line focuses on effective alignment learning with limited supervision. 
Some leverage semi-supervised learning techniques to resolve the training data insufficiency issue, including self-learning \cite{BootEA,MRAEA} and co-training \cite{KDCoE}.
(\romannumeral3) Another line of research seeks to retrieve auxiliary or indirect supervision signals from profile information or side features of entities, such as entity attributes \cite{JAPE,AttrE,MultiKE,SEA}, literals \cite{HGCN,NMN,AttrGNN}, free text \cite{JEANS}, pre-trained language models \cite{HMAN,BERTINT} or visual modalities \cite{EVA}.
Due to the large body of recent advances, we refer readers to a more comprehensive summarization in the survey \cite{OpenEA}.

\subsection{Learning with Abstention}
Learning with abstention is a fundamental machine learning, where the learner can opt to abstain from making a prediction if without enough decisive confidence \cite{BoostingAbstention,Abstention}. 
Related techniques include thresholding softmax \cite{Stefano_reject}, selective classification \cite{SelectiveClassification}, open-set classification with background classes \cite{Network_Agnostophobia} and out-of-distribution detection \cite{liang2018enhancing,vyas2018out}. The idea of learning with abstention also has applications in NLP, such as unanswerable QA, where correct answers of some questions are not stated in the given reference text \cite{SQuAD2_acl2018,AskNAQA_acl2019,ReadVerifyQA}.

To the best of our knowledge, our task, dataset, and the proposed dangling detection techniques are the first contribution to support learning with abstention for entity alignment and structured representation learning.
\section{Conclusion and Future Work}
In this paper, we propose and study a new entity alignment task with dangling cases.
We construct a dataset to support the study of the proposed problem setting, and design a multi-learning framework for both entity alignment and dangling entity detection.
Three types of dangling detection techniques are studied, which are based on nearest neighbor classification, marginal ranking, and background ranking. 
Comprehensive experiments demonstrate the effectiveness of the method, and provide insights to foster further investigation on this new problem.
We further find that dangling entity detection can, in turn, effectively provide auxiliary supervision signals to improve the performance of entity alignment.

For future work, we plan to extend the benchmarking on DBP2.0 with results from more base models of entity alignment as well as more abstention inference techniques.
Extending our framework to support more prediction tasks with abstention, such as entity type inference \cite{JOIE} 
and relation extraction \cite{alt2020tacred}, is another direction with potentially broad impact.

\section*{Acknowledgments} 
We thank the anonymous reviewers for their insightful comments. 
This work is supported by the National Natural Science Foundation of China (No. 61872172), and the Collaborative Innovation Center of Novel Software Technology \& Industrialization.
Muhao Chen's work is supported by the National Science Foundation of United States Grant IIS-2105329.

\bibliography{acl2021}
\bibliographystyle{acl_natbib}

% \clearpage
\appendix

\begin{center}
    {
    \Large\textbf{Appendices}
    }
\end{center}

\section{Degree Distribution}
\label{appendix:degree}
\begin{figure}[ht]
	\centering
	\includegraphics[width=0.8\linewidth]{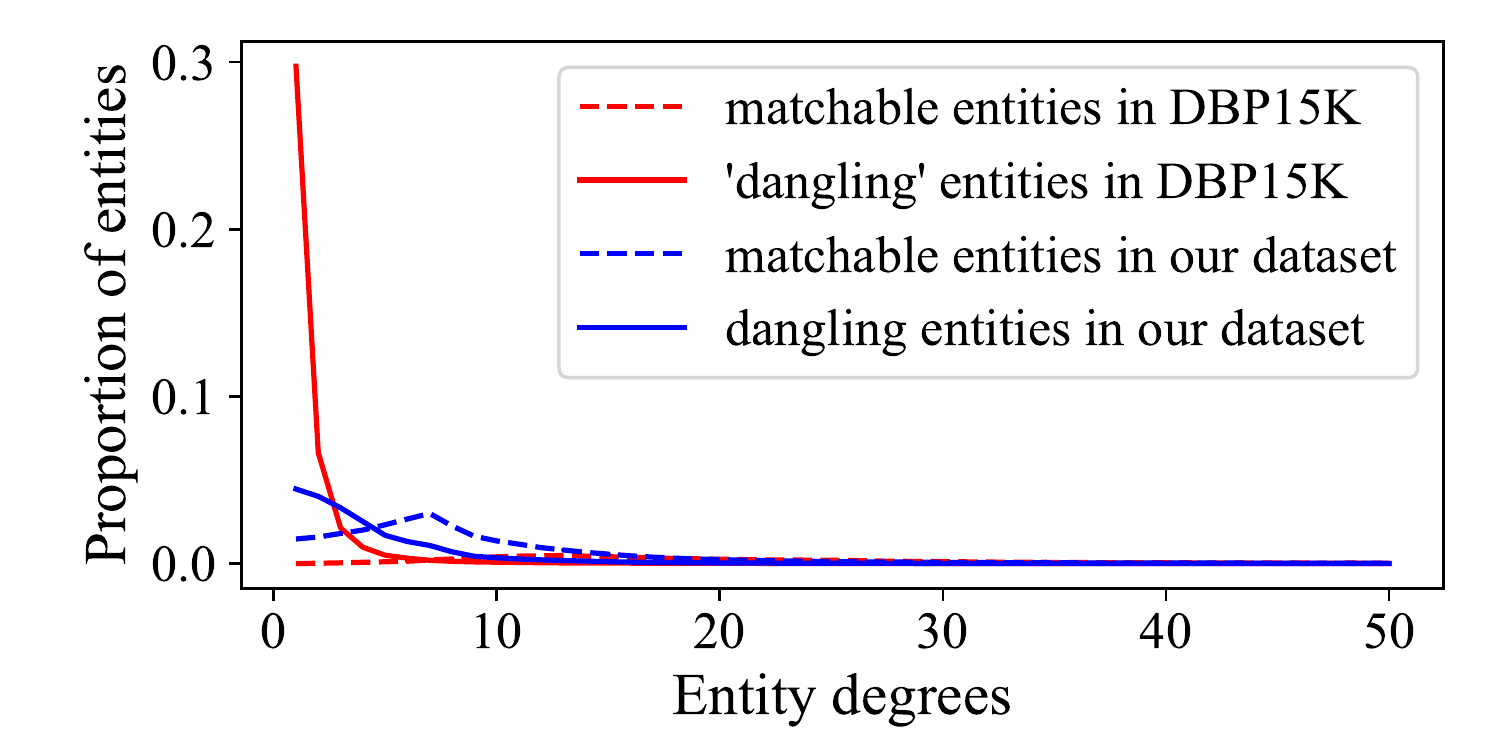}
	\caption{Degree distribution of matchable and dangling entities in DBP15K FR-EN and our FR-EN.}
	\label{fig:degree}
\end{figure}

\Cref{fig:degree} shows the degree distribution of the matchable and dangling entities in our dataset against DBP15K. 
Although DBP15K contains some entities that are not labeled to have counterparts, by checking the ILLs in the recent update of DBpedia, we find many of these entities to have counterparts in the target KG. Hence, these entities in DBP15k cannot act as dangling entities that are key to the more realistic evaluation protocol being proposed in this work.
From the comparison, we can see that these unlabeled entities in DBP15K have much fewer triples than matchable entities. 
This biased degree distribution will have an adverse effect on dangling entity detection and lead to unreal evaluation. By contrast, in our dataset, matchable and dangling entities have similar degree distribution.

\section{Configuration of MTransE and AliNet}
\label{appendix:config}
For entity alignment, we experiment with MTransE \cite{MTransE} and the SOTA method AliNet \cite{AliNet}. 
The implementation of our framework is extended based on OpenEA \cite{OpenEA}.  We adopt the truncated negative sampling method by BootEA \cite{BootEA} to generate negative triples for MTransE and negative alignment links for AliNet, which leads to improved performance. The embedding size is $128$ for MTransE and $256$ for AliNet. The batch size of MTransE is $20,480$ on ZH-EN and JA-EN, and $102,400$ on FR-EN. The batch size of AliNet is $8,192$ on ZH-EN and JA-EN, and $20,480$ on FR-EN. $\lambda_1=1.4$ in AliNet.

\section{Hyper-parameter Settings}
\label{appendix:setting}
We select each hyper-parameter setting within a wide range of values as follows:

\begin{noindlist}
  \item Learning rate: $\{0.0001, 0.0002, 0.0005, 0.001\}$
  \item Embedding dimension: $\{64, 128, 256, 512\}$
  \item Batch size: $\{4096, 8192, 10240, 20480, 102400\}$
  \item \# FNN layers: $\{1, 2, 3, 4\}$
  \item \# Random targets: $\{1, 10, 20, 30, 40, 50\}$
  \item $\lambda$: $\{0.1, 0.2, 0.3, 0.4, 0.5, 0.6, 0.7, 0.8, 0.9, 1.0\}$
\end{noindlist}

\section{Recall@10 of Entity Alignment}
\label{appendix:recall}
\begin{figure}[t]
	\centering
	\includegraphics[width=.95\linewidth]{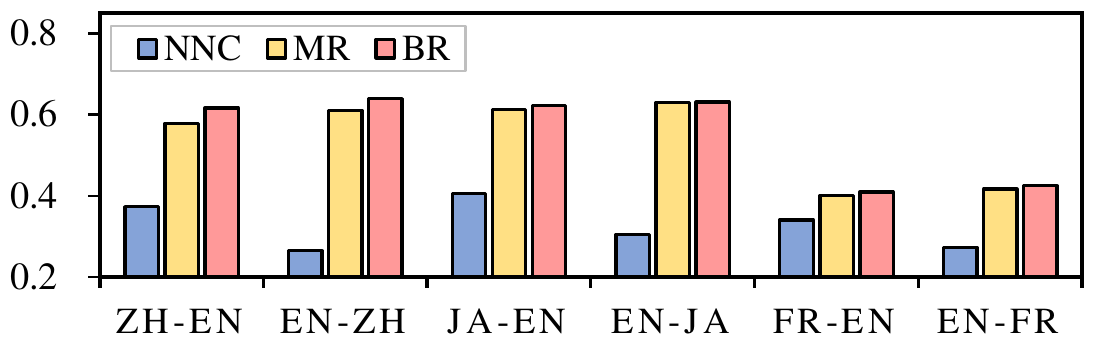}
	\caption{Recall@10 results of entity alignment.}
	\label{fig:recall}
\end{figure}

\Cref{fig:recall} gives the recall@10 results of the MTransE variants with dangling entity detection in the consolidated evaluation setting. We can see that the recall@10 results on FR-EN are lower than that on ZH-EN and JA-EN, which is similar to the observation in entity alignment~\Cref{sec:consolidated}. From the results, we think existing embedding-based entity alignment methods are still far from being usable in practice. 

\end{document}